\documentclass[12pt]{article}
\usepackage{graphicx}
\usepackage{amsmath}
\usepackage{lipsum}
\usepackage{hyperref}
\usepackage{enumitem}
\usepackage{float}
\usepackage{enumitem}

\title{\textbf{Airborne Neural Network}}
\author{
 Paritosh Ranjan \\
  IBM  \\
  \texttt{paranjan@in.ibm.com} \\
  \and
 Surajit Majumder \\
  IBM  \\
  \texttt{surajit.majumder@ibm.com} \\
  \and
 Prodip Roy \\
  IBM  \\
  \texttt{prodipro@in.ibm.com} \\
}
\date{\today}

\begin{document}

\maketitle

\begin{abstract}
Deep Learning, driven by neural networks, has led to groundbreaking advancements in Artificial Intelligence by enabling systems to learn and adapt like the human brain. These models have achieved remarkable results, particularly in data-intensive domains, supported by massive computational infrastructure. However, deploying such systems in Aerospace, where real-time data processing and ultra-low latency are critical, remains a challenge due to infrastructure limitations. This paper proposes a novel concept: the Airborne Neural Network—a distributed architecture where multiple airborne devices each host a subset of neural network neurons. These devices compute collaboratively, guided by an airborne network controller and layer-specific controllers, enabling real-time learning and inference during flight. This approach has the potential to revolutionize Aerospace applications, including airborne air traffic control, real-time weather and geographical predictions, and dynamic geospatial data processing. By enabling large-scale neural network operations in airborne environments, this work lays the foundation for the next generation of AI-powered Aerospace systems.
\end{abstract}

\section{Introduction}

Neural networks are the basic machine learning architecture behind Deep Learning models. Deep Learning is at the forefront of Artificial Intelligence systems which is solving unsolved complex problems and providing path breaking innovation due to its ability to learn on its own, just like a human brain.
More breakthroughs are expected with Neural networks as the computing infrastructural performance capacity further increases.
For example, the recent breakthrough in Generative AI is based on Deep Learning models which have been trained on huge amounts of data on enormous infrastructure.
However, if there is a need to train and run a Deep Learning system on huge compute infrastructure in Aerospace without tolerance for any delay and with lots of data being acquired continuously on the fly then currently there is no solution available.
In future, having this capability to run large deep learning systems in Aerospace can help to create innovative solutions:
\begin{enumerate}
    \item Increase the capacity of air traffic by establishing Airborne Air Traffic Control Systems which use Deep Learning models to direct each airborne vehicle
    \item Process sensor data on the fly to do new findings and more accurate and fast weather predictions
    \item Process imaging data on the fly to do new findings and more accurate geographical predictions
    \item Process geospatial data on the fly to do new findings
\end{enumerate}
Many more kinds of innovative solutions can be built if the capacity to run large neural networks with large data can be achieved in Aerospace.

\section{Brief Description of the Invention}

This invention introduces an innovative Airborne Neural Network system designed to deploy and run large-scale deep learning models across multiple airborne devices. Each device hosts a distinct group of neurons and performs computations in coordination with others via an airborne network controller and layer-specific controllers. This distributed neural network architecture enables real-time data processing, continuous learning, and inference during flight, overcoming current limitations in aerospace computing infrastructure. By leveraging the mobility and connectivity of airborne devices, the system facilitates rapid processing of sensor, imaging, and geospatial data on the fly. This breakthrough technology enables novel aerospace applications such as airborne air traffic control, improved weather forecasting, and enhanced geographical analysis, providing unprecedented AI capabilities in environments with strict latency and data volume constraints.

\section{Reduction to Practice}

Different Designs of Neuron Assembly in Airborne Devices

\subsection{Design 1}

\begin{figure}[h!]
    \centering
    \includegraphics[width=0.8\textwidth]{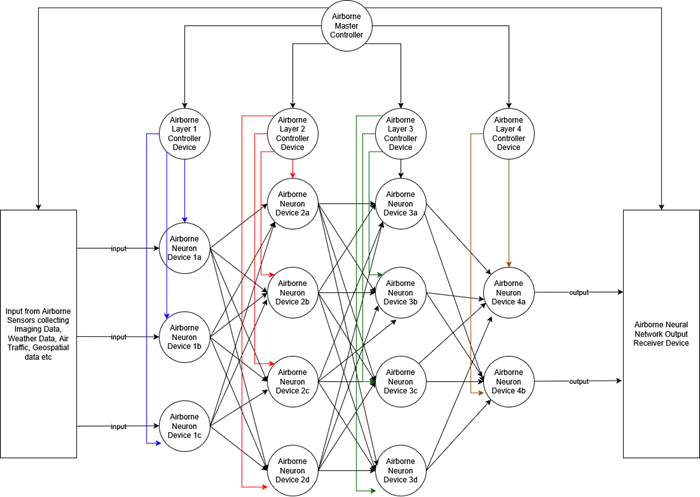}
    \caption{Airborne Neural Network}
    \label{fig:Airborne Neural Network}
\end{figure}

Each layer of a deep neural network contains one or more neurons. Each neuron or group of neurons will be in a separate airborne device in case each neuron needs that kind of server capacity to be carried airborne for compute.

\subsection{Design 2}
\begin{figure}[h!]
    \centering
    \includegraphics[width=0.8\textwidth]{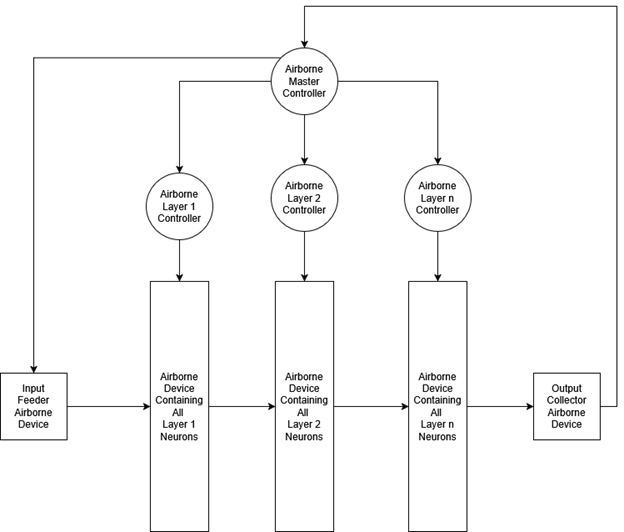}
    \caption{Airborne Neural Network carrying multiple Neurons of the same neural network layer in the same Airborne Device}
    \label{fig:Airborne Neural Network Design2}
\end{figure}
The airborne neural network will have multiple neurons of the same layer deployed in the same Airborne Device. This way even a smaller feet of airborne device could provide multi-layer airborne neural network. 

\subsection{Design 3}
\begin{figure}[h!]
    \centering
    \includegraphics[width=0.8\textwidth]{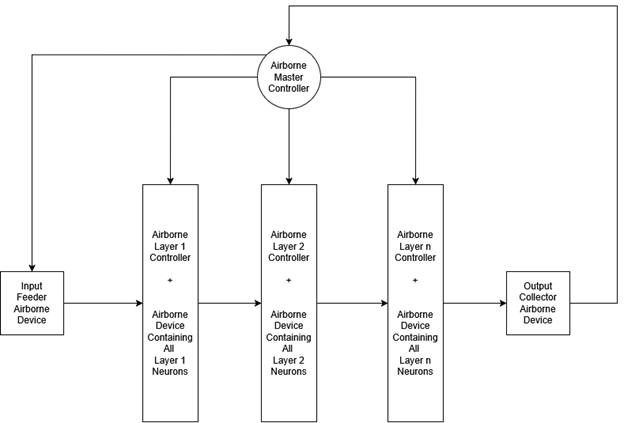}
    \caption{Airborne Neural Network carrying multiple Neurons of the same neural network layer as well as the Airborne Neural Network Layer Controller of the same layer in the same Airborne Device}
    \label{fig:Airborne Neural Network Design3}
\end{figure}
In this design, one airborne device can also have multiple neurons as well as the airborne layer controller of the same layer deployed in them. This way a even smaller fleet of Airborne Devices will be able to set up and operate an Airborne Neural Network.

The benefit of one configuration or another will depend on the use case i.e. the size of feature data, and the Neural Network algorithms used. I think the invention could facilitate any kind of structure of airborne neural network optimized either for performance or cost.

\subsection{Airborne Device Types}

Airborne devices may include various types of aerial platforms and are given below:

\begin{itemize}
    \item drone,
    \item hot air balloon,
    \item helicopter,
    \item airplane,
    \item satellite,
    \item rocket,
    \item manned aerial vehicle,
    \item unmanned aerial vehicle
\end{itemize}

Each device is capable of handling the server workload assigned to it.

\subsection{Formation}

If the devices(both neurons and sensors) are connected via wireless technologies, then any flight formation is fine if the connected layers can connect to each other and are withing range. 
However, ideally and for achieving low latency:

\begin{itemize}
    \item the devices of one layer should be behind the devices of next layer and ahead of the devices of previous layer
    \item the devices should also maintain the order of neurons for low latency.
\end{itemize}
If the devices are connected via wire, then it would be best to use the exact formation of the neural networks.

\subsection{Instructions}

Airborne Master controller would be storing and providing navigational instructions. Both manned and unmanned systems can be used. 

\section{System Components}

This system consists of the following components:

\begin{enumerate}[label=\roman*)]
    \item \textbf{Airborne Neural Network Master Controller}
    
    There is one Airborne Master Controller per Airborne Neural Network. The Airborne Master Controller contains the configuration for the entire neural network and connects with the Airborne Layer Controllers, Airborne Neural Network Input Provider, and Airborne Neural Network Output Receiver.
    
    The Airborne Master Controller is responsible for deciding and controlling:
    \begin{itemize}
        \item Navigation of the Airborne Devices
        \item Weight of each Neuron
        \item Activation Method of each Neuron
        \item Optimization Method of the Airborne Neural Network
        \item Loss function of the Neural Network
        \item Epoch – Number of Iterations for Training
        \item Network Input Data Provider
        \item Network Output Data Receiver
        \item Other Neural Network configurations
    \end{itemize}
    
    Additionally, it contains metadata of the network such as Airborne Devices, their identifiers, roles, connection methods, and authentication methods.
    
    All Airborne Devices must register with the Airborne Master Controller specifying their role. Devices are admitted into the Airborne Neural Network only after verification by the Master Controller.
    
    \item \textbf{Airborne Neural Network Layer Controllers}
    
    A deep neural network consists of multiple layers (input, output, hidden layers). Airborne Layer Controllers are airborne devices connected to the Airborne Master Controller and the Airborne Neuron Devices of their specific layer.
    
    The Layer Controller receives neuron configurations from the Master Controller and shares them with the corresponding Airborne Neuron Devices in that layer.
    
    Separate layer controllers for each layer enable scaling of the Neural Network. Multiple Layer Controllers may also exist per layer, each managing a subset of neurons to increase scalability further.
    
    \item \textbf{Airborne Neurons}
    
    Airborne Neuron Devices receive their neural network configuration from their respective Airborne Layer Controllers.
    
    Each device receives inputs from multiple neurons in the previous layer, performs computations, and sends outputs to neurons in the next layer.
    
    These devices perform key neural network operations, including calculating weighted sums and applying activation functions such as ReLU, Sigmoid, or Softmax. Based on the activation output, they determine whether to forward the data to subsequent neurons.
    
    \item \textbf{Airborne Neural Network Input Provider}
    
    These devices connect the Airborne Neural Network with various input data sources such as weather sensors, cameras, imaging data, geospatial data, air traffic control data, LiDAR, etc.
    
    They handle data preprocessing and encoding if necessary before sending the processed input to the first layer of the neural network.
    
    \item \textbf{Airborne Neural Network Output Receiver}
    
    These devices receive the network output and calculate the loss function for each training epoch or iteration.
    
    The loss data is shared with the Airborne Neural Network Master Controller, which uses optimization algorithms like Stochastic Gradient Descent or Adam (Adaptive Moment Estimation) to update neuron weights.
    
    Updated weights are then propagated back to the Layer Controllers, which configure their respective neurons for subsequent processing.
    
\end{enumerate}

\section{Low Latency}

The latency would be far less compared to sending the data to a remote data center for computing and getting the result back.
The latency can be decreased by connecting the airborne devices with loose cables will be faster than Wireless data transfer.
When the data and computing requirement become too overwhelming for the available server capacity, then distributed computing is the only way. This solution is not a perfect solution for all neural network problems. However, this invention might be able to solve problems which no other neural network can solve.

\section*{Advantages of the Invention}

Based on the invention, below are the advantages of the invention:

\begin{enumerate}
    \item \textbf{Distributed and Scalable Neural Network Processing in Aerospace}\\
    By assigning groups of neurons to dedicated airborne devices, the system enables scalable, parallel processing of large neural networks directly in the air. This distributed approach overcomes the computational limits of single devices and supports complex AI workloads in aerospace environments.
    
    \item \textbf{Real-time Data Capture and On-the-Fly Learning}\\
    The airborne neural network operates while continuously moving through aerospace, capturing real-time sensor, imaging, and signal data. This allows simultaneous data acquisition, training, and inference, enabling adaptive and timely decision-making critical for aerospace applications such as air traffic control and weather prediction.
    
    \item \textbf{Hierarchical Control Architecture for Robust Coordination}\\
    The inclusion of a master airborne device and layer-specific controllers ensures effective management, synchronization, and configuration of neurons across multiple devices. This hierarchical control enhances reliability, network integrity, and flexibility, allowing the system to adapt dynamically to changing conditions or mission requirements.
\end{enumerate}

\section{Conclusion}
This paper presents a novel Airborne Neural Network system that distributes neural network processing across multiple airborne devices. By enabling real-time data capture, training, and inference during flight, the system overcomes traditional limitations of aerospace computing infrastructure. The hierarchical control structure with a master controller and layer-specific controllers ensures robust coordination and scalability. Flexible communication links and integrated data interfaces further enhance adaptability and performance. This innovative approach opens new possibilities for advanced aerospace applications such as airborne traffic control, weather forecasting, and geospatial analysis, laying the foundation for future AI-driven airborne systems capable of operating efficiently in dynamic environments.

\section{Acknowledgment}

We would like to express our sincere gratitude to all individuals and organizations who have contributed to the success of this research. We acknowledge the invaluable support from the IBM team, whose resources and expertise have greatly enhanced this project.
Special thanks to Prodip Roy (Program Manager IBM) for their insightful feedback, guidance, and encouragement throughout the development of this work.
\section{References}
\renewcommand\refname{}

\end{document}